\documentclass[a4paper]{article}

\usepackage{INTERSPEECH2019}
\usepackage{multirow}
\usepackage{graphics}
\usepackage{verbatim}
\usepackage[inline]{enumitem}
\usepackage{xcolor}

\title{Transfer Learning from Audio-Visual Grounding to Speech Recognition}
\name{Wei-Ning Hsu, David Harwath, James Glass}
\address{Computer Science and Artificial Intelligence Laboratory \\
         Massachusetts Institute of Technology\\
         Cambridge, MA, USA}
\email{\{wnhsu,dharwath,glass\}@mit.edu}

\begin{document}

\maketitle
\begin{abstract}
Transfer learning aims to reduce the amount of data required to excel at a new task by re-using the knowledge acquired from learning other related tasks.
This paper proposes a novel transfer learning scenario, which distills robust phonetic features from grounding models that are trained to tell whether a pair of image and speech are semantically correlated, without using any textual transcripts. 
As semantics of speech are largely determined by its lexical content, grounding models learn to preserve phonetic information while disregarding uncorrelated factors, such as speaker and channel.
To study the properties of features distilled from different layers,
we use them as input separately to train multiple speech recognition models.
Empirical results demonstrate that layers closer to input retain more phonetic information, 
while following layers exhibit greater invariance to domain shift.
Moreover, while most previous studies include training data for speech recognition for feature extractor training, our grounding models are not trained on any of those data, indicating more universal applicability to new domains. 
\end{abstract}

\section{Introduction}
Robustness of automatic speech recognition (ASR) systems is essential to generalization of using speech as interfaces for human computer interaction.
Thanks to the strong modeling capacity of neural networks, recent studies~\cite{amodei2016deep, chiu2018state, jaitly2013vocal, cui2015data, ko2015audio, kim2017generation, hsu2017unsupervised, hsu2018unsupervised, hosseini2018multi, sun2018training} have demonstrated that by providing supervised examples as abundant and diverse as possible, such models can learn to extract domain invariant features and recognize linguistic units jointly.
However, without additional treatment, good performance and robustness may not be achieved when labeled data are very limited in quantities or not available in all domains~\cite{hermansky1994rasta}. 
One way to ease the burden of ASR systems is by providing better features, which are more invariant to nuisance factors while containing linguistic information ready for use (e.g., linear separability w.r.t. phonemes). 
Such features can be hand-crafted by leveraging prior knowledge~\cite{hermansky1994rasta, kingsbury1997recognizing, kim2016power, fredes2017locally}, or they can be learned in a data-driven fashion. 
Furthermore, this learning can take place jointly with ASR~\cite{sun2017unsupervised}, or separately with some tasks that have aligned objectives~\cite{hsu2017learning, hsu2018extracting, chung2019unsupervised}.

Learning features from some source tasks that can benefit the target task is a common realization of transfer learning~\cite{pan_transfer_learning_survey_2009}. 
In this work, we propose a novel inductive transfer learning scenario~\cite{pan_transfer_learning_survey_2009}, which utilizes speech features learned from audio-visual grounding for speech recognition.
Audio-visual grounding~\cite{harwath2016unsupervised} is a task which aims to distinguish whether a spoken caption is semantically associated with an image or not, and vice versa, without using any textual transcripts.
Deep audio-visual embedding network (DAVEnet)~\cite{harwath2018jointly} is a two-branched convolutional neural network model for this task, which learns to encode images and spoken captions into a shared embedding space that reflects semantic similarity.
To successfully learn a semantic representation for speech, the model has to recognize its lexical content, which in turns requires identifying phonetic content.
Therefore, one can expect intermediate layers of the speech branch in DAVEnet models to function as lexical or phonetic unit detectors.
Furthermore, since non-linguistic aspects of speech, such as speaker, are not correlated with semantics, these information may be discarded, resulting in the intermediate outputs from the model being invariant to domain shift.

We conduct a series of ASR experiments probing properties of the features distilled from DAVEnet models at different layers.
Results indicate higher in-domain accuracy using features closer to input, and better robustness to domain shift using features from latter layers.
In addition, we also study how the choice of DAVEnet architectures and grounding performance affects the performance of distilled feature extractors.
In summary, our contributions are three-fold:
\begin{enumerate*}[label=(\arabic*)]
    \item To the best of our knowledge, this is the first work connecting audio-visual grounding with speech recognition.
    \item Our empirical study verifies that the distilled feature extractors not only contain sufficient information for recognizing phonemes, but better remove nuisance information.
    \item Moreover, the grounding models are trained on a different dataset from that used for ASR, indicating more general applicability of the distilled features.
\end{enumerate*}

\section{Learning Spoken Languages through Audio-Visual Grounding}
In this section, we describe in detail the source task as well as the DAVEnet model, and then review several analysis studies which lay the foundation for our work. 

\begin{figure}
    \centering
    \includegraphics[width=\linewidth]{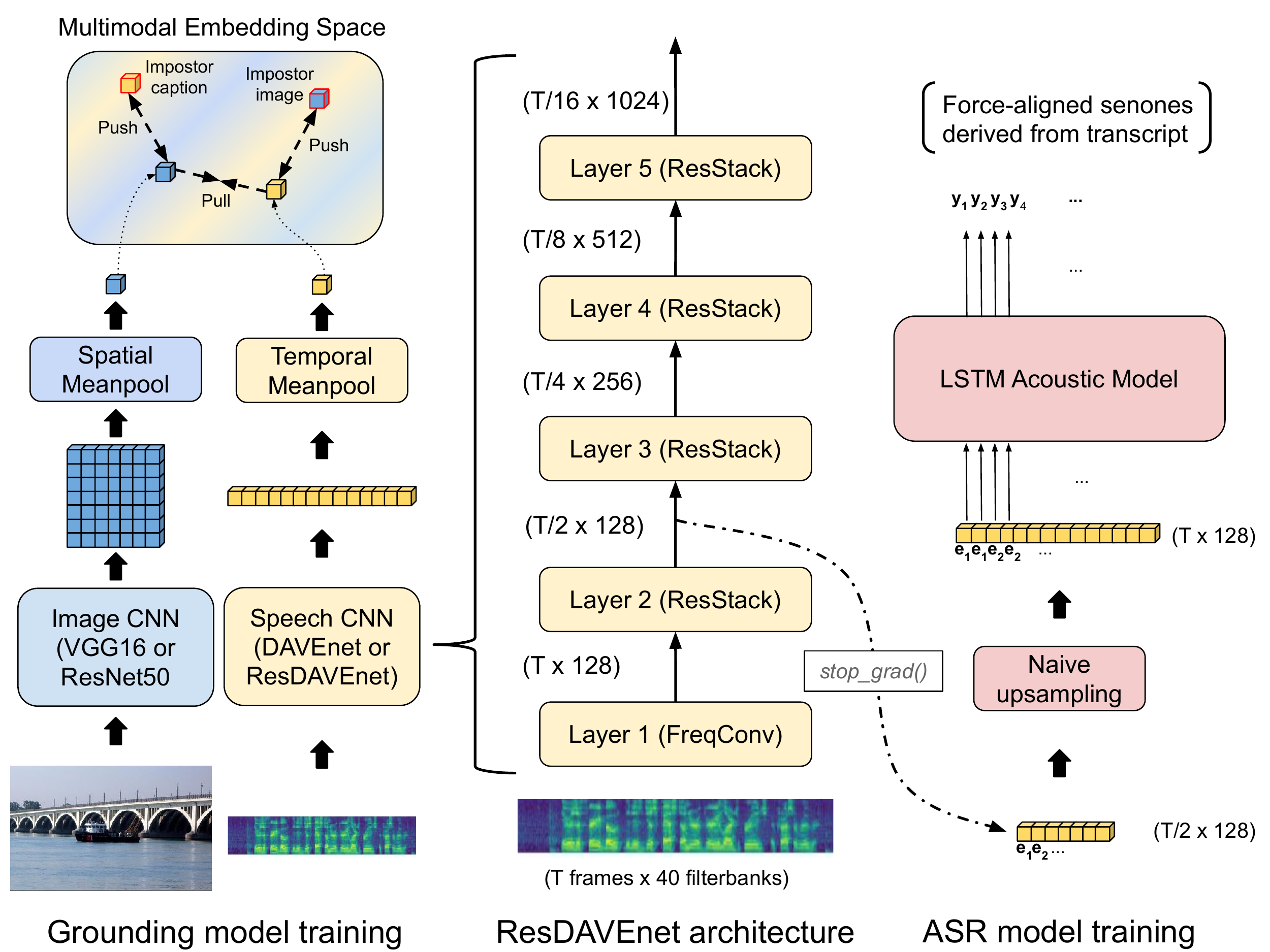}
    \caption{Graphical illustration of audio-visual grounding model training (left), ResDAVEnet architecture (center), and feature distillation pipeline for speech recognition (right).}
    \label{fig:model}
\end{figure}

\subsection{Audio-Visual Grounding}
Inspired by the fact that humans learn to speak before being able to read or write, 
audio-visual grounding of speech is a proxy task proposed in~\cite{harwath2016unsupervised} that aims to examine the capability of computational models to learn a language using only semantic-level supervision.
To simulate such a learning scenario, a model has access to images and their spoken captions during training.
The goal of the model is to learn a semantic representation for each caption and each image, 
such that representations of semantically correlated utterances and images are similar to each other, while those from irrelevant pairs are dissimilar.
Performance is evaluated using a cross-modality retrieval task: given a spoken sentence, a model is asked to rank a list of 1,000 images according to semantic relevance, with only one image being the correct answer, and vice versa.
Recall@10 averaged over the retrieval tasks in both directions is used for evaluation.

\subsection{Deep Audio-Visual Embedding Network (DAVEnet)}
DAVEnet is a convolutional neural network (CNN) for audio-visual grounding proposed in~\cite{harwath2016unsupervised, harwath2017learning, harwath2018jointly}, which consists of two branches: $f$ for speech and $g$ for image, as depicted in Figure~\ref{fig:model}.
Each branch has a sequence of strided convolutional blocks, followed by a global mean-pooling layer to produce a fixed dimensional representation.
The model is trained to minimize a triplet loss~\cite{hoffer2015deep, jansen2018unsupervised}: given a similarity measure $S(\cdot, \cdot)$, paired speech and image, $x_s$ and $x_i$, along with one imposter instance from each modality, $\tilde{x}_s$ and $\tilde{x}_i$, the loss enforces $S(f(x_s), g(x_i))$ to exceed both $S(f(x_s), g(\tilde{x}_i))$ and $S(f(\tilde{x}_s), g(x_i))$ by a predefined margin.
Following~\cite{harwath2019towards}, imposter instances are drawn using a mixture of uniform sampling and within-batch semi-hard negative mining~\cite{jansen2018unsupervised}.
$S(z_1, z_2) = \langle z_1, z_2 \rangle $ is used here.

In our experiments, we make use of two DAVEnet model variants. 
The first is identical to the model used in \cite{harwath2018jointly}, which uses an audio network comprised of 5 convolutional layers and the VGG16 architecture for the image network. 
The second model, ResDAVEnet, is based upon deep residual networks \cite{resnet}. 
The image network makes use of the ResNet50 architecture, while the audio network is based on strided 1-D convolutions with residual connections. 
The first layer of the ResDAVEnet audio model is comprised of 128 convolutional units each spanning all frequency channels but only one temporal frame, with a temporal shift of 1 frame. 
This is followed by a ReLU and a BatchNorm layer. 
The remainder of the network is a sequence of 4 residual stacks with channel dimensions 128, 256, 512, and 1024. 
Each residual stack is comprised of a sequence of two basic residual blocks (as described in \cite{resnet}) which share the same overall channel dimension, with 2-D 3x3 kernels replaced with 1-D kernels of length 9.
Additionally, the first residual block in each layer in each stack is applied with a stride of two frames, resulting in an effective temporal downsampling ratio of $2^4$ for the entire network, as shown in Figure~\ref{fig:model} (center).

\subsection{Emergence of Multi-Level Speech Unit Detectors}
Recent work~\cite{drexler2017analysis, harwath2017learning, harwath2019towards} on analyzing DAVEnets have shown that, despite the fact that phoneme and word labels are never explicitly provided, such detectors automatically emerge within these models.
Phoneme-like detectors reside in layers closer to the input~\cite{drexler2017analysis, harwath2019towards}, while semantic word detectors reside in layers closer to the output~\cite{harwath2017learning}.
Such findings echo with the recent discovery in the computer vision community~\cite{zhou2014learning, zhou2015object} that in a trained scene classifier, layers closer to sensory input appear to be low-level pattern  (e.g., shape, edge, and color) detectors, while object detectors emerge at later layers.
This behavior can be mainly attributed to the compositionality of the prediction target as well as the inductive bias we impose in the model architecture.
Just as a scene can often be determined by the objects that are present, the semantics of a spoken sentence is determined by the sequence of words, each of which in turn is determined by phoneme sequences.
Prediction of semantic objects from a spoken sentence can therefore be regarded as a bottom-up process, which iteratively composes higher-level concepts from lower-level ones with the hierarchical convolution operations in CNNs.

\section{Transfer Learning to Speech Recognition}

\subsection{Distilling Robust Feature Extractors for ASR}
Both DAVEnet variants are trained on the Places Audio Caption dataset (PlacesAudCap)~\cite{harwath2018jointly}, derived from the Places205 scene classification dataset~\cite{zhou2014learning}.
PlacesAudCap is composed of over 400K image and unscripted spoken caption pairs collected from 2,954 speakers via Amazon Mechanical Turk, which sums up to over 1,000 hours.
For the audio-visual grounding task, both models use 40-dimensional log Mel filterbank (FBank) features with 10ms shift and 25ms analysis window as input, and achieve R@10 of 0.629 and 0.720, respectively.

As a natural result of large-scale crowd-sourcing, this dataset exhibits great diversity not only in textual content, but also in speaker, background noise, and microphone channels.
For both semantic grounding and speech recognition, these non-textual factors are nuisances to the target, and therefore would eventually be removed from the internal representations learned by the networks trained for these two tasks.
Having been exposed to a vast amount of nuisance factors, we hypothesize that the audio branch of DAVEnet models would also learn domain invariant phonetic representations at later layers, which can be subsequently utilized for robust speech recognition.
From now on, we denote features extracted from the $k$-th layer of model $M$ with $M$-L$k$, for example, ResDAVEnet-L2.

To account for the different frame rates at different layers in DAVEnet models, when extracting outputs from a layer with a down-sampling rate $r$ compared to the speech inputs, we repeat each step $r$ times for simplicity, as shown in Figure~\ref{fig:model} (right).

\subsection{Evaluating Transfer Learning Performance}
To evaluate transfer learning performance, we consider three criteria:
\begin{enumerate*}[label=(\arabic*)]
    \item inclusion of phonetic content, 
    \item exclusion of nuisance factors, and
    \item transferrability across datasets.
\end{enumerate*} The first two are evaluated using a protocol similar to~\cite{hsu2018extracting}, where an ASR model is trained on a set of domains, and evaluated on both in-domain and out-of-domain speech (relative to the training data).
Performance on in-domain data characterizes an upper bound for the amount of phonetic information that can be inferred from the input.
The performance gap between in-domain and out-of-domain data quantifies the invariance of the features to nuisance factors: 
the smaller this gap, the more invariant the features are.
To test the third criteria, instead of training the source task on a dataset that includes speech used for the target task, a separate dataset collected through a different process (i.e., PlacesAudCap) is used. 
We emphasize here that this is a more practical setting to consider than training one feature extractor for each target task.

\section{Related Work}

Transfer learning has a long history in the field of machine learning~\cite{pan_transfer_learning_survey_2009}. 
More recently, deep neural network models have been shown to be extremely effective for learning representations of data with a high degree of re-usability across many different tasks and domains. 
Perhaps the most well-known example of this is the use of the ImageNet~\cite{imagenet} image classification database to pre-train convolutional neural network models for other downstream computer vision tasks~\cite{Razavian_2014,faster_rcnn,two_stream_cnns}. 
Other sub-fields have also developed similarly techniques. 
For example, in natural language processing, dense word vector models such as word2vec~\cite{word2vec} and GloVe~\cite{glove}, or more advanced ones like ELMo~\cite{elmo} and BERT~\cite{bert} have quickly replaced one-hot word representations in many tasks and pushed the state-of-the-art forward on a variety of language understanding tasks.
More recently, there is also an increasing interest in learning from multimodal data~\cite{yeh2018unsupervised} and transfer learned representations from such tasks~\cite{gupta2017aligned}

In the field of speech recognition, low-resource speech recognition is a scenario which heavily benefits from transfer learning, for example in the form of training on multilingual datasets~\cite{ekapol_2016}.
Other models capable of disentangling phonetic and domain information have recently been shown to learn acoustic features with a greater degree of domain invariance than traditional acoustic features~\cite{hsu2017learning, hsu2017unsupervised, hsu2018extracting}. 
Another line of work has studied the use of the visual modality as a form of weak supervision using semantic information for acoustic modeling~\cite{harwath2016unsupervised,kamper_taslp19,chrupala_2017}, followed up with analysis on representations learned from such models~\cite{drexler2017analysis,alishahi_2017,harwath2019towards}.
In this paper, we build upon this prior work and quantify the degree to which these representations can be used to build robust ASR.

\section{Experiments}

\subsection{ASR Setup and Baselines}
We consider TIMIT~\cite{zue1990speech} and Aurora-4~\cite{pearce2002aurora} for training ASR systems to study robustness of the proposed method to speaker, channel, and noise.
TIMIT contains 5.4 hours of 16kHz broadband recordings of read speech from 630 speakers, of which about 70\% are male.
Recordings from male speakers are used for training ASR systems, which are then tested on both genders.
Aurora-4 is based on the Wall Street Journal (WSJ) corpus~\cite{garofalo2007csr}, containing recordings with microphone and noise variation.
The set of conditions are divided into four groups: clean (A), noisy (B), channel (C), and noisy+channel (D).
While recordings in A are recorded by one microphone in quiet environments, those in C are recorded with a different set of microphones than A.
Recordings in B and D are created from A and C, respectively, with artificially added noises.
Similar to~\cite{hsu2018extracting}, we use the clean set (A) for training ASR systems, and test on the four groups separately.

Kaldi~\cite{povey2011kaldi} is used for training of initial HMM-GMM models, forced alignment, and decoding. 
The Microsoft Cognitive Toolkit (CNTK)~\cite{seide2016cntk} is used for neural network-based acoustic model training.
To simplify the pipeline and study only the effect of ASR input features, the same forced alignment derived from a HMM-GMM model trained on Mel-frequency cepstral coefficient (MFCC) features are used for all experiments, following the default recipe in Kaldi.
A three-layer long short-term memory (LSTM) acoustic model with 1,024 memory cells and a 512-node linear projection at each layer is used~\cite{sak2014long}. 
Training of LSTM acoustic models closely follows~\cite{zhang2016highway}, which minimizes a frame-level cross-entropy loss using stochastic gradient descent with a momentum of 0.9 starting from the second epoch.
Initial learning rate is set to 0.2 per minibatch, and $L2$ regularization with a weight of $1e-6$ is used.

We consider two types of features to compare with our proposed method.
The first one is FBank feature, which is the input to DAVEnet models and contains rich phonetic and domain information.
The second one is the latent segment variable $z_1$ from a model called factorized hierarchical variational autoencoder (FHVAE)~\cite{hsu2017learning}.
FHVAE learns to encode sequence-level and segment-level information into separate latent variables without supervision by optimizing an evidence lower bound derived from a factorized graphical model, and has been shown effective for extracting domain invariant ASR features~\cite{hsu2018extracting}.

While previous work investigated usage of FHVAE for ASR by training FHVAE models on all domains of the target task (e.g., Aurora-4 with all four conditions)~\cite{hsu2018extracting, hsu2018unsupervised}, 
we also evaluate FHVAE models trained on PlacesAudCap to test cross-dataset transferability, and on the subset of domains used for ASR training.
We use FHVAE models with two LSTM layers, each with 256 cells, for both the encoders and decoder. 
A discriminative weight of $\alpha=10$ is applied for all models, and the scalable training algorithm proposed in~\cite{hsu2018scalable} is used for training on PlacesAudCap dataset with a sequence batch size $K=5000$, because the original algorithm cannot handle large-scale datasets.

\subsection{Main Results}
Tables~\ref{tbl:a4_res} and \ref{tbl:timit_res} present the testing word error rates (WERs) on both in-domain and out-of-domain conditions for ASR systems trained with different features. 
\textit{FE Train Set} denotes the data used for training feature extractors, and \textit{A/I} following \textit{Places} represents the audio and image portion of the PlacesAudCap dataset, respectively.

Starting with Table~\ref{tbl:a4_res}, we observe that FBank suffers from severe degradation in all out-of-domain conditions (B, C, and D), while FHVAE trained on all conditions of the Aurora-4 dataset achieves the best performance.
However, when trained on \textit{Places A}, improvement of FHVAE from FBank on out-of-domain data becomes less significant in the presence of additive noise, compared to the result in the purely channel-mismatched condition (C).
Results of the proposed methods are shown in the second and the third section in Table~\ref{tbl:a4_res}.
While features from ResDAVEnet consistently outperforms FBank and FHVAE (Places A) for all layers, those from DAVEnet do not.
We hypothesize that the much deeper architecture of ResDAVEnet at each layer (ResStack) enables better removal of nuisance factors and preserving of linguistic information compared to DAVEnet, which also reflects in the comparison of grounding performance as mentioned earlier.

It is also worth noting that, for both DAVEnet and ResDAVEnet models, performance in matched domain degrades when using latter layers, and except for ResDAVEnet-L1, all features are actually worse than the FBank baseline.
This could indicate discarding of relevant phonetic information in the process of inferring higher-level semantic representation such as words.
Table~\ref{tbl:timit_res} demonstrates a similar trend as Table~\ref{tbl:a4_res}, where FHVAE trained on TIMIT dataset of all genders achieves the best out-of-domain WER, and ResDAVEnet-L2 is the best comparing to models trained on Places. 

\begin{figure}[t]
    \centering
    \includegraphics[width=\linewidth]{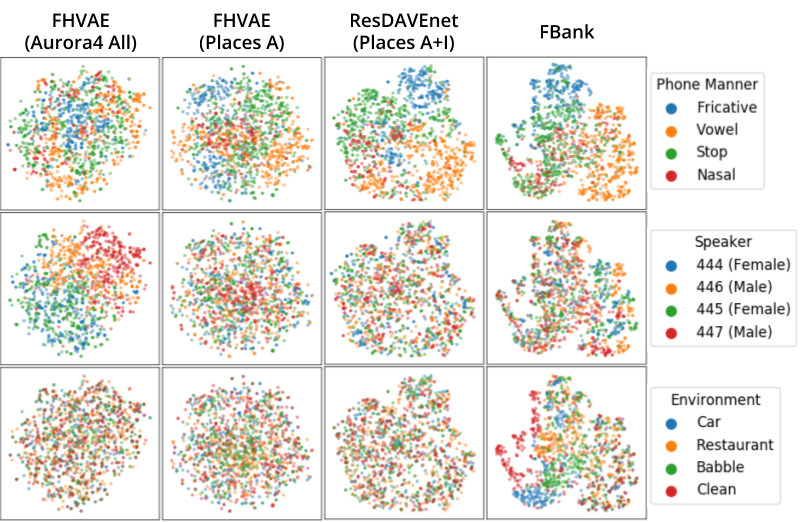}
    \caption{Frame-level t-SNE projections for four different acoustic representations, color coded for phonetic manner class, speaker identity, and noise/environment type. Visually, the ResDAVEnet features encode the least amount of speaker and environment information.}
    \label{fig:aurora4_tsne}
    \vspace{-1.5ex}
\end{figure}

We also present qualitative visualizations in Figure \ref{fig:aurora4_tsne} using t-SNE~\cite{tsne} comparing ResDAVEnet, FHVAE (Places A / Aurora4 All), and the baseline FBank feature. 
It can be observed from the first row that all three features contain phonetic information, as different phonetic manners are separated in each space.
On the other hand, the project features of ResDAVEnet and FHVAE (Aurora4 All) are visually more environment invariant than those from the other two (for the FHVAE trained on PlacesAudCap, green and orange dots concentrate more at the center than red and blue dots).
Such visualization correlates well with the performance of the various feature types in Tables~\ref{tbl:a4_res}.

To conclude, we learn that
\begin{enumerate*}[label=(\arabic*)]
    \item despite being trained with exactly the same process, inductive bias introduced to model architectures (i.e., DAVEnet versus ResDAVEnet) still affects the properties of learned representations, 
    \item feature extractors distilled from ResDAVEnet models clearly preserve phonetic information while improving invariance to nuisance factors, and most importantly,
    \item it achieves better cross-dataset transferrability compared to FHVAE and FBank features.
\end{enumerate*}

\begin{table}[t]
\begin{center}
\caption{\label{tbl:a4_res} Aurora-4 test WERs for different ASR features. A is the domain matched with the ASR training set.}
\resizebox{\linewidth}{!}{%
\begin{tabular}{lllllll}
\multirow{2}{*}{ASR Feature} & \multirow{2}{*}{FE Train Set} & \multicolumn{5}{c}{Test WER (\%)} \\
& & Avg. & A & B & C & D\\
\hline\hline
FBank           & N/A           & 53.38 & 4.02  & 50.77 & 40.13 & 66.31 \\
FHVAE           & Places A      & 49.31 & 4.37  & 44.43 & 26.64 & 65.33 \\
\hline
DAVEnet-L1      & Places A+I    & 57.89 & 3.40  & 54.92 & 46.89 & 71.69 \\
DAVEnet-L2      & Places A+I    & 57.05 & 4.56  & 56.15 & 34.88 & 70.35 \\
DAVEnet-L3      & Places A+I    & 61.65 & 8.52  & 60.53 & 35.57 & 75.90 \\
\hline
ResDAVEnet-L1   & Places A+I    & 44.03 & \textbf{2.91}  & 38.53 & 36.86 & 57.53 \\
ResDAVEnet-L2   & Places A+I    & \textbf{33.11} & 4.20  & \textbf{25.17} & 27.09 & 46.75 \\
ResDAVEnet-L3   & Places A+I    & 33.16 & 7.23  & 25.24 & \textbf{26.38} & \textbf{46.46} \\
ResDAVEnet-L4   & Places A+I    & 42.76 & 15.02 & 36.38 & 32.43 & 55.45 \\
\hline\hline
FHVAE           & Aurora4 (Clean)   & 71.98 & 4.75  & 72.54 & 50.57 & 86.15 \\
FHVAE           & Aurora4 (All)     & 24.41 & 5.01  & 16.42 & 20.29 & 36.33
\end{tabular}}
\vspace{-4.0ex}
\end{center}
\end{table}

\subsection{Correlation with Source Task Performance}
Finally, we study how the performance of the grounding task affects the transfer learning performance, conditioning on the same neural network architecture for the source task.
We create two proper subsets of 200k and 80k paired image/audio captions, and train one ResDAVEnet model on each subset.
R@10 of the retrieval task for the models trained with 80k, 200k, and 400k (original) are 0.343, 0.582, and 0.720, respectively.

Results are shown in Table~\ref{tbl:a4_wer_frac}.
Except for the first layer, we can observe that WER decreases as the amount of source task training data increases.
In fact, except for the out-of-domain conditions of the first layer, all layers improve in all conditions (full results not shown due to space limit). 
Discovery of such positive correlation affirms the relatedness of the two tasks and encourages collection of a larger dataset for building a general feature extractor based on semantic grounding tasks.

\begin{table}[t]
\begin{center}
\footnotesize
\caption{\label{tbl:timit_res} TIMIT test WERs by gender for different ASR features.}
\resizebox{0.7\linewidth}{!}{%
\begin{tabular}{llll}
\multirow{2}{*}{ASR Feature} & \multirow{2}{*}{FE Train Set} & \multicolumn{2}{c}{Test WER (\%)} \\
& & Male & Female \\
\hline\hline
FBank           & N/A           & \textbf{20.39} & 31.15 \\
FHVAE           & Places A      & 25.35 & 33.22 \\
\hline
DAVEnet-L1      & Places A+I    & 20.58 & 30.62 \\
DAVEnet-L2      & Places A+I    & 21.94 & 32.57 \\
DAVEnet-L3      & Places A+I    & 28.64 & 32.74 \\
\hline
ResDAVEnet-L1   & Places A+I    &  21.48 & 30.74 \\
ResDAVEnet-L2   & Places A+I    & 22.28 & \textbf{27.40} \\
ResDAVEnet-L3   & Places A+I    & 27.26 & 29.31 \\
ResDAVEnet-L4   & Places A+I    & 38.60 & 42.07 \\
\hline\hline
FHVAE           & TIMIT (All)   & 22.00 & 26.20
\end{tabular}
}
\vspace{-3.2ex}
\end{center}
\end{table}

\begin{table}[t]
\begin{center}
\caption{\label{tbl:a4_wer_frac} Aurora-4 average test WERs for using features extracted from ResDAVEnet trained on different sizes.}
\footnotesize
\resizebox{0.8\linewidth}{!}{%
\begin{tabular}{lllll}
FE Train Set & L1 & L2 & L3 & L4 \\
\hline\hline
Places A+I (80k)    & \textbf{41.69} & 39.52 & 43.42 & 51.45 \\
Places A+I (200k)   & 43.46 & 37.50 & 37.85 & 44.18 \\
Places A+I (400k)   & 44.03 & \textbf{33.11} & \textbf{33.16} & \textbf{42.76} \\
\end{tabular}
}
\vspace{-6ex}
\end{center}
\end{table}

\section{Concluding Discussion and Future Work}
In this paper, we present a successful example of transfer learning from a weakly supervised semantic grounding task to robust ASR.  We achieve cross-dataset transferability, which is an important milestone toward building a generalized feature extractor to be used in many tasks and domains like BERT.
In addition, along with the analysis in~\cite{drexler2017analysis, harwath2019towards}, this work sheds light on using semantic level supervision to learn the compositional structure of a language.
For future work, we would like to study methods for leveraging target task data, possibly through semi-supervised training or adaptation, in order to bridge the gap to FHVAE trained on those data.
Furthermore, unlike FHVAE, it is unclear at which layer a ResDAVEnet model learns to maximally remove domain information. 
We would also like to explicitly force such disentanglement to occur at certain layers, which can possibly improve both the grounding performance and the robustness of distilled features.

\bibliographystyle{IEEEtran}
\bibliography{main}

\end{document}